\pdfoutput=1
\documentclass[11pt]{article}

\usepackage[final]{acl}
\usepackage{times}
\usepackage{latexsym}

\usepackage[T1]{fontenc}
\usepackage[utf8]{inputenc}

\usepackage{microtype}

\usepackage{inconsolata}
\usepackage{booktabs} % 导入booktabs宏包以使用三线表
\usepackage{multirow}
\usepackage{graphicx}
\usepackage{graphicx}
\usepackage[table]{xcolor}
\usepackage{amsmath,amssymb,amsfonts}
\usepackage{algorithmic}
\usepackage{textcomp}
\usepackage{xcolor}
\usepackage{hyperref} % 超链接
\usepackage{url}      % simple URL 

\usepackage[normalem]{ulem} % 控制下划线间距 

\usepackage{multirow}
\usepackage{multicol}
\usepackage{array}
\usepackage{rotating} 
\usepackage{tabularx} % 调整表格大小，自适应layout
\usepackage{makecell} % \Xhline命令是由makecell包提供的，用来绘制较粗的水平线，可以在表格中使用
\usepackage{setspace} % \setstretch
\usepackage{tcolorbox} % 文本换行高亮
\usepackage{enumitem} % 更灵活地控制列表的样式和缩进；
\usepackage{caption} % 调整caption与前后之间的距离
\usepackage{booktabs}
\usepackage[table]{xcolor} % 支持表格背景色
\usepackage{multirow}
\usepackage{caption}

\definecolor{lightgray}{gray}{0.9} % 定义浅灰色

\usepackage{algorithm} 


\definecolor{lzqPurple}{RGB}{148, 39, 212}
\definecolor{lightRed}{RGB}{253, 190, 172}
\definecolor{lightGreen}{RGB}{172, 253, 187}
\definecolor{lightYellow}{RGB}{253, 248, 172}
\title{
ART: Attention Replacement Technique to Improve Factuality in LLMs
}

\author{
 \textbf{Ziqin Luo\textsuperscript{2}}*,
 \textbf{Yihao Quan\textsuperscript{1}}*,
 \textbf{Xiaofeng Zhang\textsuperscript{1}*$^\dagger$},
 \textbf{Xiaosong Yuan\textsuperscript{3}}, \textbf{Chen Shen\textsuperscript{3}}
\\
 \textsuperscript{1}Department of Automation and Intelligent Sensing, Shanghai Jiao Tong University \\
 \textsuperscript{2}Fudan University 
 \textsuperscript{3}Alibaba Cloud Computing \\
    {\tt\small \{framebreak@\}sjtu.edu.cn}\quad
}
\begin{document}
\maketitle
\begin{abstract}
% This document is a supplement to the general instructions for *ACL authors. It contains instructions for using the \LaTeX{} style files for ACL conferences.
% The document itself conforms to its own specifications, and is therefore an example of what your manuscript should look like.
% These instructions should be used both for papers submitted for review and for final versions of accepted papers.
\begingroup
\renewcommand\thefootnote{} % 临时清空编号内容
\footnotetext{${}^*$ These authors contributed equally to this work}
\footnotetext{${}^\dagger$ Corresponding author}
\endgroup
Hallucination in large language models (LLMs) continues to be a significant issue, particularly in tasks like question answering, where models often generate plausible yet incorrect or irrelevant information. Although various methods have been proposed to mitigate hallucinations, the relationship between attention patterns and hallucinations has not been fully explored. In this paper, we analyze the distribution of attention scores across each layer and attention head of LLMs, revealing a common and intriguing phenomenon: Shallow layers of LLMs primarily rely on uniform attention patterns, where the model distributes its attention evenly across the entire sequence. This uniform attention pattern can lead to hallucinations, as the model fails to focus on the most relevant information. To mitigate this issue, we propose a training-free method called \textcolor{red}{\textbf{A}}ttention \textcolor{red}{\textbf{R}}eplacement \textcolor{red}{\textbf{T}}echnique (ART), which replaces these uniform attention patterns in the shallow layers with local attention patterns. This change directs the model to focus more on the relevant contexts, thus reducing hallucinations. Through extensive experiments, ART demonstrates significant reductions in hallucinations across multiple LLM architectures, proving its effectiveness and generalizability without requiring fine-tuning or additional training data.

\end{abstract}

\section{Introduction}
\label{sec:introduction}

\begin{figure}[h]
\centering % 图片居中
\captionsetup{aboveskip=0pt, belowskip=0pt} % 仅对该figure的caption设置前后间距
\includegraphics[width=1.0\linewidth,page=1]{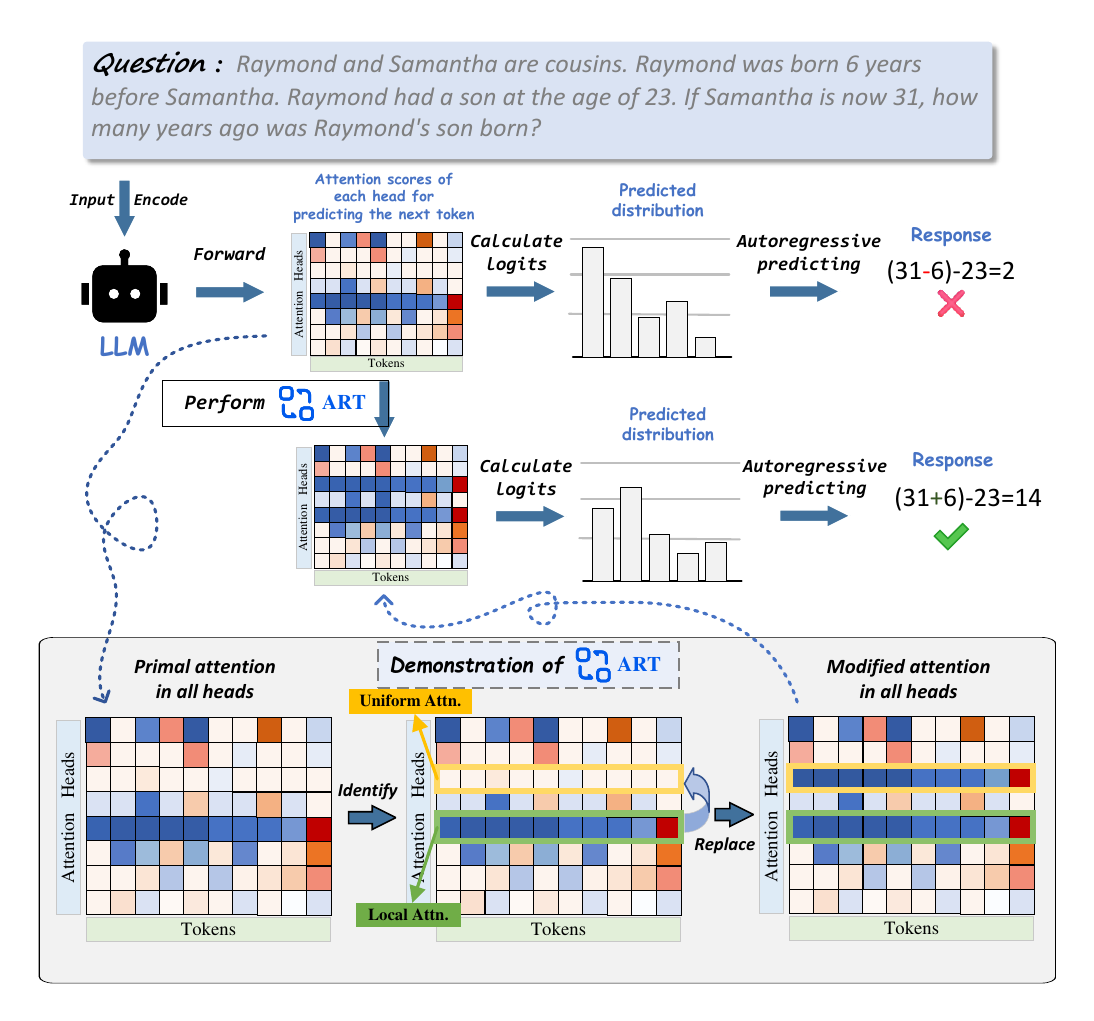}
% \vspace{0.01cm}
\caption{
% The structure diagram of ART replaces the uniform attention head in each attention head that predicts the next token with a local attention head. Baseline: \textit{Llama2-7B-Chat} processes a sample from TruthfulQA.%~\cite{22-truthful-qa}. 
A demonstration overview of how ART works during the decoding process, with $N_h=8$ and $k=1$.
}
% \label{motivation1}
\label{art_intro}
\end{figure}

\begin{figure}[h]
\centering % 图片居中
\captionsetup{aboveskip=0pt, belowskip=0pt} % 仅对该figure的caption设置前后间距
\includegraphics[width=1.0\linewidth,page=1]{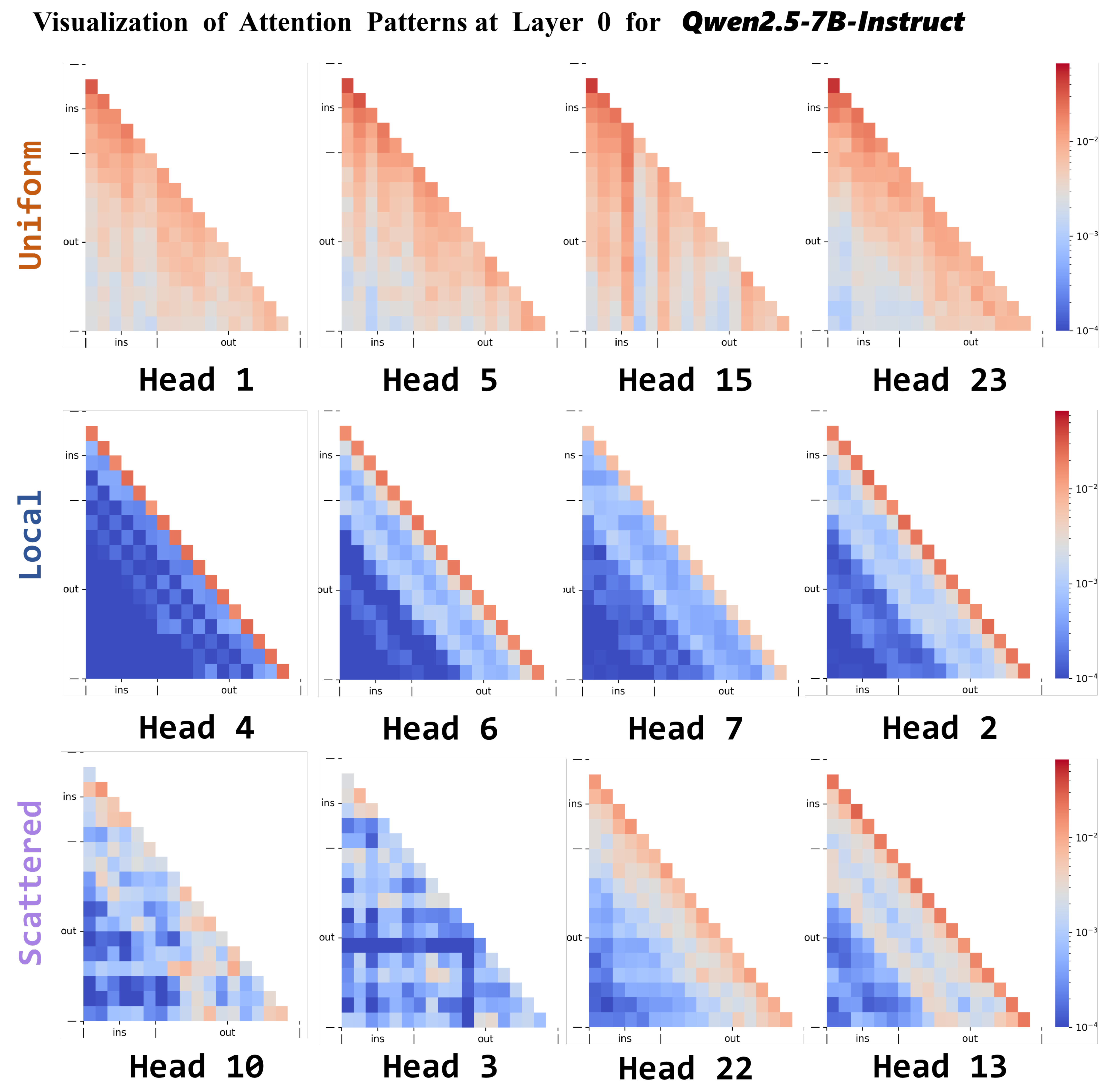}
% \vspace{0.01cm}
\caption{
Visualization on some of attention weights of \textbf{Layer 0} in \textit{Qwen2.5-7B-Instruct} model when encoding a sample from GSM8K.%~\cite{21-gsm8k}. 
Attention weights are divided into 3 categories: \textit{Uniform Attention}, \textit{Local Attention}, and \textit{Scattered Attention}.
}
\label{fig:attention-patterns}
\end{figure}

%原 Large language models (LMMs) have made significant strides, while hallucinations remain a persistent challenge, particularly in tasks such as Question Answering (QA). Current methods for addressing hallucinations often involve changing decoding strategies, incorporating external knowledge bases, or retraining models with additional data \cite{24-llama-slayer}. These approaches, however, often require significant resources and time.
Large language models (LLMs) have made significant strides, but hallucinations remain a persistent challenge, particularly in tasks such as Question Answering (QA). This is a highly challenging issue for the practical deployment of large models. Therefore, to ensure that the model's output is true and reliable, existing research attempts to solve the hallucination problem by incorporating external knowledge bases or retraining models with additional data \cite{24-llama-slayer}. However, these approaches often require significant resources and may substantially reduce the inference efficiency.

% Recent research into attention sink has offered new insights into hallucinations. The concept of attention sink as an information flow is introduced in ``Label Words are Anchors" \cite{label-words}, which shows how information flow often converges on a specific user token in large language models (LLMs). OPERA and DOPRA \cite{opera,dopra} further explores the connection between attention sink in user tokens and output tokens in MLLMs. It observes that when a token has a high attention weight across subsequent tokens, this over-reliance on the token can lead to hallucinations in the model's outputs. Although these methods clarify the relationship between attention sink, user tokens, and output tokens, the relationship between attention sink, image tokens, and hallucination remains unclear. It's important to note that MLLM's output tokens are generated by the decoder based on logits, whereas input tokens, which constitute most of the input sequence, are more likely to directly reflect the MLLM's internal mechanisms.

% 原Recent work such as Selayer8B \cite{SLayer8B} suggests that models intervene mainly on the first two sides because subsequent layers are sparse, i.e., the model focuses mainly on contextual information, as shown in Fig. \ref{fig:attention-patterns}, where we visualize this sparseness in llama to illustrate the situation of the attention graphs of the shallow and deep layers.
To strike a trade-off between performance and resource consumption, researchers have intervened in the attention mechanism of transformer-based large language models to mitigate hallucinations \cite{zhou2025mitigating,opera, eah,shi2025meaningless,shi2026improving,zhang2025cot,zhang10,zhang2,zhang8,zhang9,zhang2025shallow,zhang2025enhancing}. This empirical study provides valuable insights and reveals the important role that attention heads play in alleviating hallucinations. More broadly, recent studies have also examined consistency and reliability issues in adjacent multimodal and agent settings \cite{quan2026reinforcing,zhang2025dive}.

In addition to head-level analysis, there are also layer-level analyses. Recent research \cite{SLayer8B,zhang2025redundancy,che2026counting} reveals that shallow layers, particularly the first two layers, play a more critical role in knowledge injection and deserve denser injection, while deeper interventions have minimal effect and could even be pruned. AdaInfer \cite{fan2024not} also supports this idea, suggesting that current LLMs can perform adequately if truncated at certain intermediate layers, as the middle and later layers do not contribute much more to the intermediate representations in some tasks.

Combining head-level and layer-level analyses and ideas, we believe that shallow-layer attention heads play an important role in mitigating hallucinations. Therefore, we visualize the attention maps of the various attention heads in shallow layers, as shown in Figure~\ref{fig:attention-patterns}. These attention heads can be categorized into three types based on their distribution: local, uniform, and medium. The characteristics of these types are as follows: (1) \textbf{Uniform heads} distribute the model's attention evenly across the context. (2) \textbf{Local heads} only attends to neighboring tokens, and the visualized results are relatively sparse. (3) \textbf{Scattered heads} focuses predominantly on certain preceding tokens.
% fig:attention-patterns

Since layers are not fully decoupled, a combination of most layers must be executed sequentially to achieve the best results \cite{sun2024transformer}. ITI \cite{23-iti} also indicates that the local pattern is crucial for the model’s understanding of semantics. Therefore, we aim to maximize the effectiveness of shallow layers in understanding semantics and contextual information, thus mitigating hallucinations while maintaining the integrity of the model.

Motivated by these observations, we propose an attention-enhancement method in shallow layers, which replaces uniform attention heads with local attention heads to enhance the information exchange capability of the model's attention, and thus improve the model’s performance. We conduct extensive evaluations, specifically focusing on hallucination issues, and test mainstream LLMs to validate the effectiveness of ART in reducing hallucinations across various model architectures. Our results demonstrate that ART is a highly effective plug-and-play solution for mitigating hallucinations in various LLMs.

Specifically, our contributions can be summarized as follows:
\begin{itemize}[leftmargin=0.5cm, topsep=-2pt] % 调整左侧缩进宽度和减少 itemize 环境请后的垂直距离
% \begin{itemize} 
\item We systematically studied the distribution characteristics of shallow-layer attention heads, categorizing them into three types: Uniform, Local, and Scattered. Through experiments, we validated the crucial role of local attention in the model's generation process.
% \vspace{-1.0em}
\item To address the hallucination problem in large language models (LLMs), we propose a training-free method called Attention Replacement Technique (ART) to replaces redundant uniform attention heads with local attention heads in the shallow layers, significantly improving the truthfulness of the model's output.
% \vspace{-2em}
\item Experiments on multiple models validate the plug-and-play convenience and strong generalization of this method.
\end{itemize}
\section{Related Work}
% \subsection{Attention sink and information flow}
\textbf{Attention sink and information flow}.
StreamingLLM \cite{24-streaming-llm} observes high attention values on the first token, termed an "attention sink," and leverages this finding to extend the input sequence length—a positive use of high attention values.
However, different scenarios can exhibit different behaviors.   Several studies have demonstrated the negative effects of the attention sink phenomenon.
The ACT \cite{24-act} study found that attention sinks not only occur on the first token but also on tokens with limited semantic information (e.g., ".", ":", and "<0x0A>") . Contrary to the observations made in StreamingLLM—which suggest preserving attention sinks to enhance LLMs' accuracy—they highlight that not all attention sinks are beneficial.   Specifically, for most attention sinks occurring in the middle or later parts of inputs, reducing their attention scores can lead to improved accuracy.
In OPERA \cite{opera} and DOPRA \cite{dopra}, it was found that when models generate hallucinated content, the self-attention weights in the last layer exhibit a distinct "columnar" pattern before the hallucination occurs, leading to an "over-trust" tendency in the self-attention weights for the hallucinated parts.
% Yuan et al \cite{yuan} proposed Instance-adaptive prompting to select better prompt to LLMs for correct reasoning for different instances, and IAP-ss compared the significance scores and thresholds of the information flow to determine the appropriate prompt.
\citet{yuan} proposed Instance-adaptive prompting to select better prompt to LLMs for correct reasoning for different instances, and IAP-ss compared the significance scores and thresholds of the information flow to determine the appropriate prompt.
EAH \cite{eah} reveals a phenomenon: most hallucinations are closely related to the attentional sinking pattern in the image-labeled self-attention matrix, where shallower layers show intensive attentional sinking and deeper layers show sparse attentional receptions. They propose a training-free approach called Enhanced Attention Head (EAH) designed to enhance image convergence by focusing attention on shallow layers.

% \subsection{Attention Heads of Large Language Models}
\textbf{Attention Heads of Large Language Models}.
LoFiT \cite{lofit}'s extends the task from model realism more generally, for a particular downstream task, the first step is to first identify the subset of attentional heads that are most important for the task by learning, and then the second step is to train a corresponding offset vectors for each of the attentional heads in this subset of heads, with the core purpose is to optimize the activations of these heads.
\citet{sixteen} argues that the nlp model, even if trained in the mode of multi-head attention, can remove a large fraction of these attention heads during real-world testing without affecting the model's performance.
\citet{retrieval} argues that all models with long context capabilities have a set of retrieval heads, and pruning out the retrieval heads will result in the loss of the model's ability to retrieve relevant information and create illusions, whereas pruning the non-retrieval heads does not affect the model's retrieval ability.
\citet{duoattention} argues that the attention heads within the LLM can be roughly divided into two parts: retrieval heads and streaming heads; the former is important for the model to find key information in context, and cropping the (contextual) KV cache within retrieval heads significantly affects the final output of the language model, while modifying the KV cache of streaming heads has no The final output of the language model is significantly affected by cropping the (contextual) KV cache within retrieval heads while modifying the KV cache of streaming heads does not.

\begin{figure*}[h]
\centering % 图片居中
\captionsetup{aboveskip=0pt, belowskip=0pt} % 仅对该figure的caption设置前后间距
\includegraphics[width=0.8\linewidth,page=1]{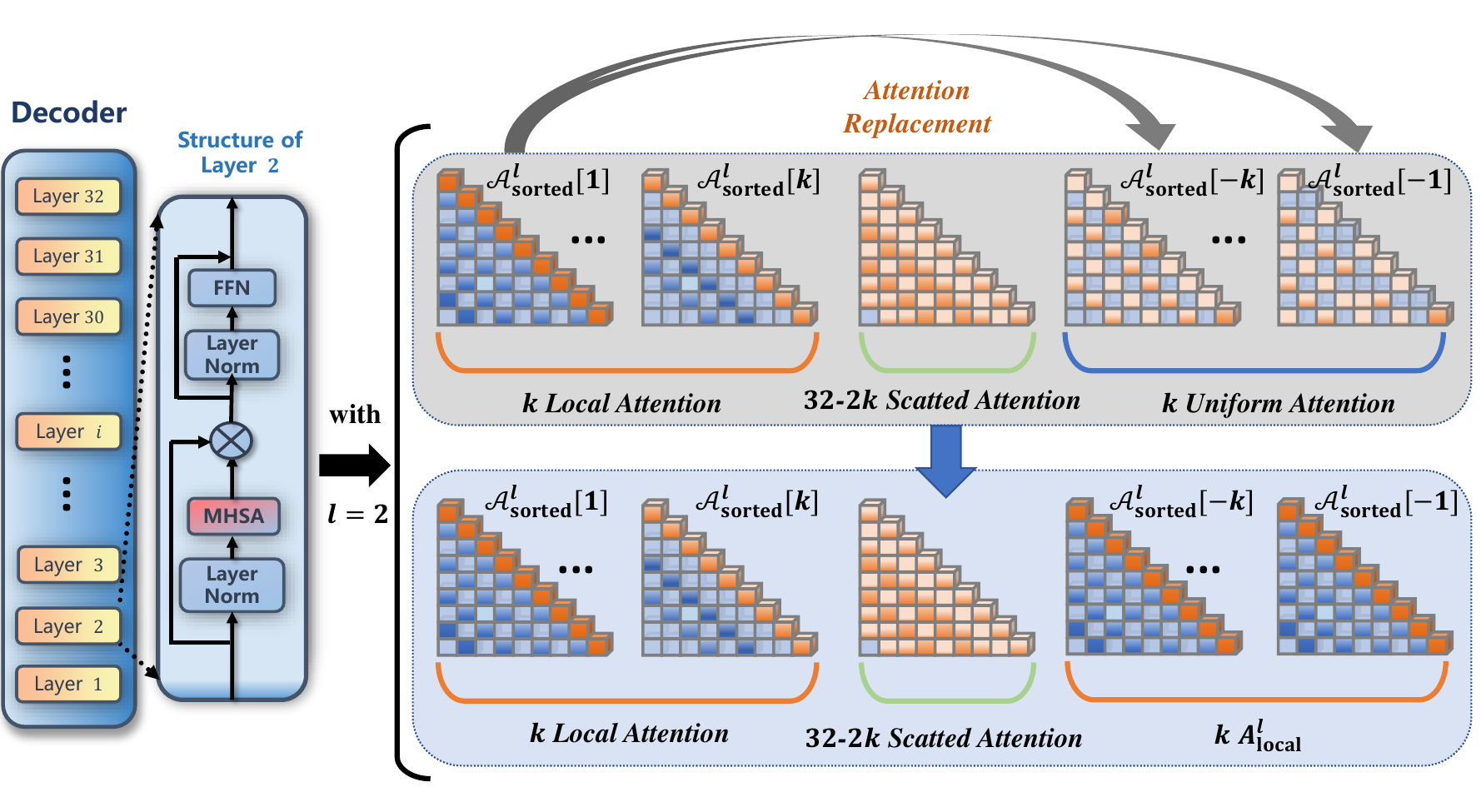}
% \vspace{-0.1cm}
\caption{
% The visualization of the averaged attention weights across all heads for each layer when the language model processes a sample from TruthfulQA~\cite{22-truthful-qa}. 
% Cases presented are from \textit{Llama2-7B-Chat} and \textit{Qwen2.5-7B-Instruct}.
The detailed demonstration of applying \text{ART-max} to the second layer of \textit{Llama2-7B-Chat}.
}
\label{fig:sparse-dense-attention}
% \vspace{-0.05cm}
\end{figure*}

\section{Preliminaries}
\label{sec:preliminaries}
% 参考 when attention sinks emerges in language models
% 介绍llm中的注意力机制
% since流行的大语言模型(gpt4,llama,qwen,mistral)主要以 decoder-only 结构，在本文中我们所有讨论的大语言模型默认是 decoder-only结构。大语言模型由L个解码层堆叠而成，每一个解码层包含两个子层，分别称为MHA和FFN，他们共同编码和理解输入的T个输入标记并以串行的方式连接。每个解码层中还包含层归一化模块用以调激活值的整数值分布。

% This paper only discusses LLMs built with the decoder-only structure since prevalent LLMs~\cite{24-llama3,24-qwen2.5} are mainly based on this structure.
A decoder-only LLM is composed of $L$ stacked decoder layers.
Each layer consists of two modules: Multi-Head Attention (MHA) and Feed-Forward Network (FFN).
They are connected serially and encode the $T$ input tokens $\boldsymbol{t}=\{t_1, t_2, \dots, t_T\}$ jointly.
Formally,  Let the input of the $l$-th layer be  $\boldsymbol{H}^l = \{\boldsymbol{h}^l_1, \boldsymbol{h}^l_2, \dots, \boldsymbol{h}^l_T\} \in \mathbb{R}^{T\times d}$.
The $t$-th token is encoded with a densen hidden state $\boldsymbol{h}^{l}_{t}$ with dimension being $d$.
% Since existing LLMs mainly adopt a pre-norm setting, 我们将第l个解码层表示为 pre-norm 的格式：
% Since LLMs mainly adopt a pre-norm structure, we represent the $l$-th layer in that format:
Since LLMs mainly adopt a pre-norm structure, we represent the $l$-th layer:
\begin{align}
    \boldsymbol{O}^l &= \text{MHA}^{l}\left(\text{LN}(\boldsymbol{H}^l)\right) \\
    \boldsymbol{H}^{l+1} = &\text{FFN}^{l}(\text{LN}(\boldsymbol{O}^l+\boldsymbol{H}^l)) + \boldsymbol{O}^l + \boldsymbol{H}^l.
\end{align}
% \in\mathbb{R}^{T\times d}
% $\text{MHA}^{l}$ is the $l$-th layer's multi-head attention operator, projecting the input sequence $\boldsymbol{X}\in \mathbb{R}^{T\times d}$ into $N_h$ subspaces to perform multiple attention mechanisms in parallel and aggregating them at last:
$\text{MHA}^{l}$ is the $l$-th layer's multi-head attention operator, projecting the input sequence $\boldsymbol{X}\in \mathbb{R}^{T\times d}$ into $N_h$ subspaces to perform attention mechanisms in parallel and aggregating them at last:
\begin{align}
    \label{eq:multihead-attention}
    \footnotesize\text{MHA}^{l}(\boldsymbol{X}) &= \footnotesize\sum_{h=1}^{N_h}\text{\small{Softmax}}\left( \frac{\boldsymbol{Q}_h^{l}(\boldsymbol{K}_h^{l})^\top}{\sqrt{d_h} }\right)\boldsymbol{V}_h^{l}\boldsymbol{W}_{o,h}^{l} \\
    \label{eq:x_to_qkv}
    \footnotesize
    \boldsymbol{Q}_h^{l} = \boldsymbol{X}&
    \footnotesize\boldsymbol{W}_{q,h}^{l}, 
    \boldsymbol{K}_h^{l} = \boldsymbol{X}\boldsymbol{W}_{k,h}^{l}, 
    \boldsymbol{V}_h^{l} = \boldsymbol{X}\boldsymbol{W}_{v,h}^{l}. 
\end{align}
Here, $\boldsymbol{W}_{q,h}^{l}, \boldsymbol{W}_{k,h}^{l}, \boldsymbol{W}_{v,h}^{l}\in\mathbb{R}^{d\times d_h}$ and $\boldsymbol{W}_{o,h}^{l}\in\mathbb{R}^{d_h\times d}$ are parametric matrices belonging to the $h$-th head of $l$-th layer. 
$d_h=d/N_h$ represents the dimension of each projected head.
To investigate how each token in the sequence attends itself and its previous tokens, we denote the attention weight of the $h$-th head as 
\begin{equation}
    \boldsymbol{A}_{h}^{l} \equiv \text{Softmax}\left( \frac{\boldsymbol{Q}_h^{l}(\boldsymbol{K}_h^{l})^\top}{\sqrt{d_h} }\right).
\end{equation}
Actually, attention weights vary among different heads, capturing different patterns in the input sequences~\cite{17-transformer}.
Thus, MHA makes language models more expressive than vanilla attention.
In the remainder of this paper, unless otherwise specified, the concept of \textit{attention} refers to the attention weight $\boldsymbol{A}_{h}^{l}$.

\subsection{Attention Patterns in LLMs}
\label{sec:attention-patterns}
% 参考 ACT 的方式进行写作
% 主要介绍的内容： 在我们的标准下llms中有哪些attention pattern，每种 pattern 的特点（定义，位于llm哪里，长啥样-对应怎样的模式，它对于llm生成有何作用？） 
% overview-总起
% 先讨论 llm 中的 attention 类型：按照深度来分类：分为 dense attention 和 sparse attention （用图来说明）；然后将讨论引导到浅层的 dense attention；对于浅层的 dense attention，我们大致可以将其分类3类：uniform attention、local attention 和 scattered attention；（解释我们的分类的定义和划分逻辑）；
% 本节需要用图片来加以辅助说明
% 核心思路：1. 不同层的注意力重要性不同；2. 不同层的注意力模式不同；=> 直接结果，对不同的 layer 

% 近期许多研究(attention-survey, streamingllm, act, fastv)对大语言模型在处理输入序列时的内部注意力头进行了广泛的调查(investigate)并对处于不同层的不同注意力进行了详细的可视化。
Recent studies~\cite{24-attention-survey,24-streaming-llm, 24-act, 24-fastv} have investigated the internal attention heads of LLMs as they process input sequences, providing detailed visualizations of attentions across different layers.
% 这些研究表明大语言模型的注意力具有丰富的模式。
These studies reveal that attentions in LLMs exhibit diverse patterns.
% 我们目标是调查不同的注意力并探索它们对大语言模型生成过程的影响。(这句借鉴ACT) 为了实现这一目标，我们根据相关研究以及实践结果对不同的注意力模式进行分类 (section 4.1)，并探索不同注意力模式对模型生成的影响(section 4.2)。
We aim to investigate these different attention patterns and explore their influences on LLMs' generation. 
To achieve this goal, we categorize attention patterns~(Section~\ref{sec:attention-category}) and examine their effects~(Section~\ref{sec:attention-influence}).

\begin{figure}[h]
\centering % 图片居中
\captionsetup{aboveskip=0pt, belowskip=0pt} % 仅对该figure的caption设置前后间距
\includegraphics[width=1.0\linewidth,page=1]{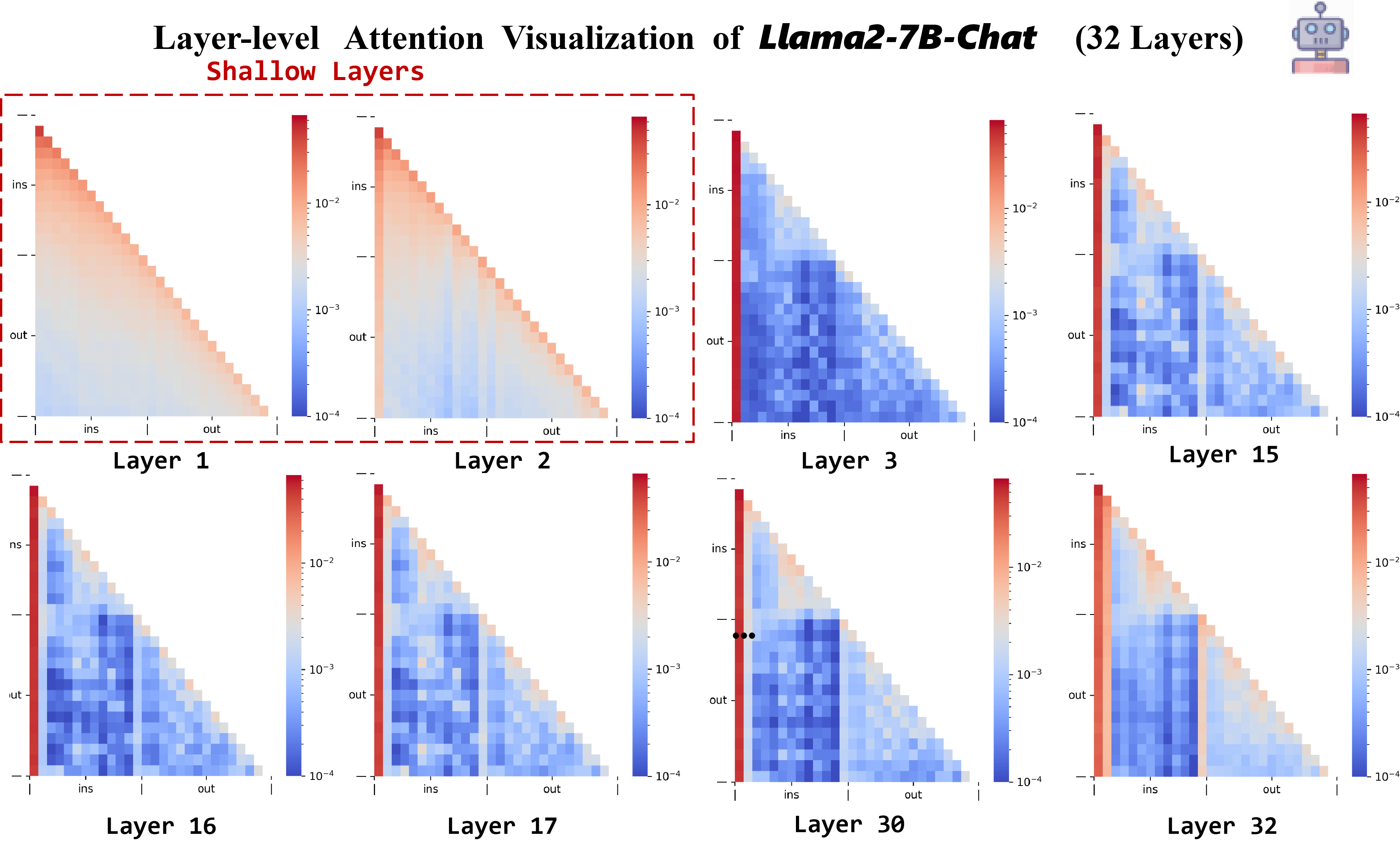}
% \includegraphics[width=1.0\linewidth,page=1]{latex/figures/art_intro.pdf}
% \vspace{0.01cm}
\caption{
The structure diagram of ART replaces the uniform attention head in each attention head that predicts the next token with a local attention head. Baseline: \textit{Llama2-7B-Chat} processes a sample from TruthfulQA.%~\cite{22-truthful-qa}. 
}
\label{motivation1}
\end{figure}

\subsection{Category}
\label{sec:attention-category}
% ！！ 这里要展示图来直观说明 dense attn 和 sparse attn； ！！（分别展示 shallow layer 和 deep layer 的平均注意力）
\textbf{Sparse and dense attention}. 
% \lzq{Add Figures to demonstrate dens attn and sparse attn...}
% ACT 和 FastV 对大语言模型内的各层注意力进行了可视化。我们可以根据是否存在注意力池(引用streamingLLM)来将注意力模式划分稀疏注意力和密集注意力。具体而言，注意力池标记会从其他可以访问它的标记处吸收大量的注意力分数，导致其他标记所获取的注意力分数很少，从而导致稀疏注意力。而当不存在注意力池标记时，各个标记所获取的注意力分数相对均衡，此时对应密集注意力。
Yu et al.~\cite{24-act} and Chen et al.~\cite{24-fastv} visualize multiple attentions across various layers within LLMs.
We can categorize attention patterns into \textit{sparse attention} and \textit{dense attention} based on the presence of attention sinks~\cite{24-streaming-llm}. 
Specifically, attention sinks are tokens that absorb a significant amount of attention scores from other tokens that can attend them, resulting in sparse attention patterns where the attention scores obtained by other tokens are dramatically low. 
In contrast, when attention sinks are absent, the attention scores are more evenly distributed among tokens, corresponding to dense attention.
% 如图xx所示，从视觉上看，稀疏注意力中大部分标记获取的注意力接近于0，整体颜色偏深。密集注意力中注意力分布相对密集和均衡，整体颜色偏浅。
% As shown in Figure~\ref{fig:sparse-dense-attention}, 
As shown in Figure~\ref{motivation1}, 
visually, in sparse attention, most tokens receive nearly zero attention, resulting in a generally darker appearance. 
In dense attention, the attention distribution is relatively balanced, leading to a lighter overall color.
% 一般而言，稀疏注意力主要出现在llm的深层，而密集注意力主要出现在llm的浅层。
Generally, sparse attention predominantly occurs in LLMs’ deeper layers, while dense attention appears primarily in the shallower layers. 
% 值得注意的是，Chen等人(SLayer)表明模型的shallow layers相对于deep layers对知识注入和理解更重要。FastV的实验也表明编辑shallow layer相对于编辑deep layer 对模型的生成影响更大。因此，本文聚焦于对大语言模型浅层注意力的讨论，以求尽可能对注意力进行轻量化干预。遵循(following)其他工作(进行相关的引用)的设定，本文将浅层定义为模型的前两层。
In particular, Chen et al.~\cite{24-llama-slayer} demonstrate that shallow layers are more important for the injection and understanding of knowledge compared to deeper layers.
FastV's experiments~\cite{24-fastv} also indicate that editing shallow layers has a greater impact on the generation process than editing deep layers. 
Therefore, this paper discusses attentions in the shallow layers, aiming to intervene in attention as minimally as possible. 
Following the settings of previous studies~\cite{24-fastv,24-act}, unless otherwise specified, we define shallow layers as the first two layers of LLMs.

\textbf{Three patterns for dense attention}.
% ！！ 此外除了平均注意力，还需展示单个 head 的注意力，以说明 dense attention 里面还可以继续划分 pattern； —— 最左边是 uniform, 最右边是 local, 中间放几张 中等 ranking 的 attention plot！！
% \lzq{add figures to illustrate 3 attn patterns...}
% 对于在浅层的密集注意力，不同注意力头所展现的注意力模式也不同。如图xxx所示，密集注意力大致可以分为三种模式：uniform, local, and scattered.
Different attention heads exhibit different attention patterns in shallow layers of dense attention. 
As illustrated in Figure~\ref{fig:attention-patterns}, dense attention can generally be categorized into three patterns: uniform, local, and scattered.
% 它们代表了三种不同的注意力分布策略。
These represent three different characteristics of attention distribution.
% 均匀注意力中每个token都近似平等地attend所有前文标记；局部注意力中每个token只attend邻近标记；散点注意力中，每个token会将注意力着重分配到前文中的某些标记上。
In uniform attention, each token attends almost equally to preceding tokens; in local attention, each token only attends to neighboring tokens; in scattered attention, each token focuses predominantly on certain preceding tokens. 
% 我们通过定义一个叫做m指数的指标来划分这三类注意力。形式上来说，令xxx (写得数学一点)
For better illustration, we define the $m$-index $m_h^l$ to classify these three types of attention.
Formally, let $\boldsymbol{U}\in\mathbb{R}^{T\times T}$ denotes the completely uniform attention 
\begin{align}
\boldsymbol{U}[i,j] = \left\{\begin{matrix}
% 0 &  i>j\\
% 1/i &  i\le j\\
0 &  i<j\\
1/i &  i\ge j\\
\end{matrix}\right.,
\end{align}
where $\boldsymbol{U}[i,j]$ is the attention weight between the $i$-th and $j$-th tokens in input tokens $\boldsymbol{t}$. We define $\boldsymbol{A}_{h}^{l}$ as the attention of the $h$-th head in the $l$-th layer and $m_{h}^{l}$ as:
\begin{align}
\label{eq:m-index}
\footnotesize
m_{h}^{l} = \frac{1}{N(N+1)/2}\sum_{i\ge j}\max\left(\frac{\boldsymbol{A}_{h}^{l}[i,j]}{\boldsymbol{U}_{h}^{l}[i,j]}, \frac{\boldsymbol{U}_{h}^{l}[i,j]}{\boldsymbol{A}_{h}^{l}[i,j]}\right) 
\in\mathbb{R}^{+}.
\end{align}
The smaller $m_{h}^{l}$, the more similar $\boldsymbol{A}_{h}^{l}$ is to $\boldsymbol{U}_{h}^{l}$, corresponding to uniform attention. 
In contrast, a larger value of $m_{h}^{l}$ signifies a greater distinction between $\boldsymbol{A}_{h}^{l}$ and $\boldsymbol{U}_{h}^{l}$, which refers to local attention.
% 描述不同模式注意力对应的指标情况。
The transition from uniform attention to local attention can be viewed as a continuous spectrum characterized by their corresponding $m_{h}^{l}$. 
Scattered attention represents an intermediate state along this spectrum.
% 由于不同T会影响m指数的绝对值，所以m指数的相对大小更具有意义。具体来说，对于同一层的所有注意力头，计算它们各自的m指数并按照m指数的大小进行排序。
Since the absolute value of the $m$-index is influenced by the sequence length $T$, the relative magnitude of different $m$-indices is more meaningful. 
Specifically, for all attention heads within the same layer, we calculate their respective $m$-indices and rank these indices by magnitude.
% 排序最靠前的注意力视为均匀注意力，排序最末尾的注意力视为局部注意力，位于排序中间的注意力则可视为散点注意力。
Attentions ranked at the top are considered uniform attention, while those at the bottom are regarded as local attention. 
The rest in the middle of the ranking are considered scattered attention.

\begin{figure}[t]
\centering % 图片居中
\captionsetup{aboveskip=0pt, belowskip=0pt} % 仅对该figure的caption设置前后间距
\includegraphics[width=1.0\linewidth,page=1]{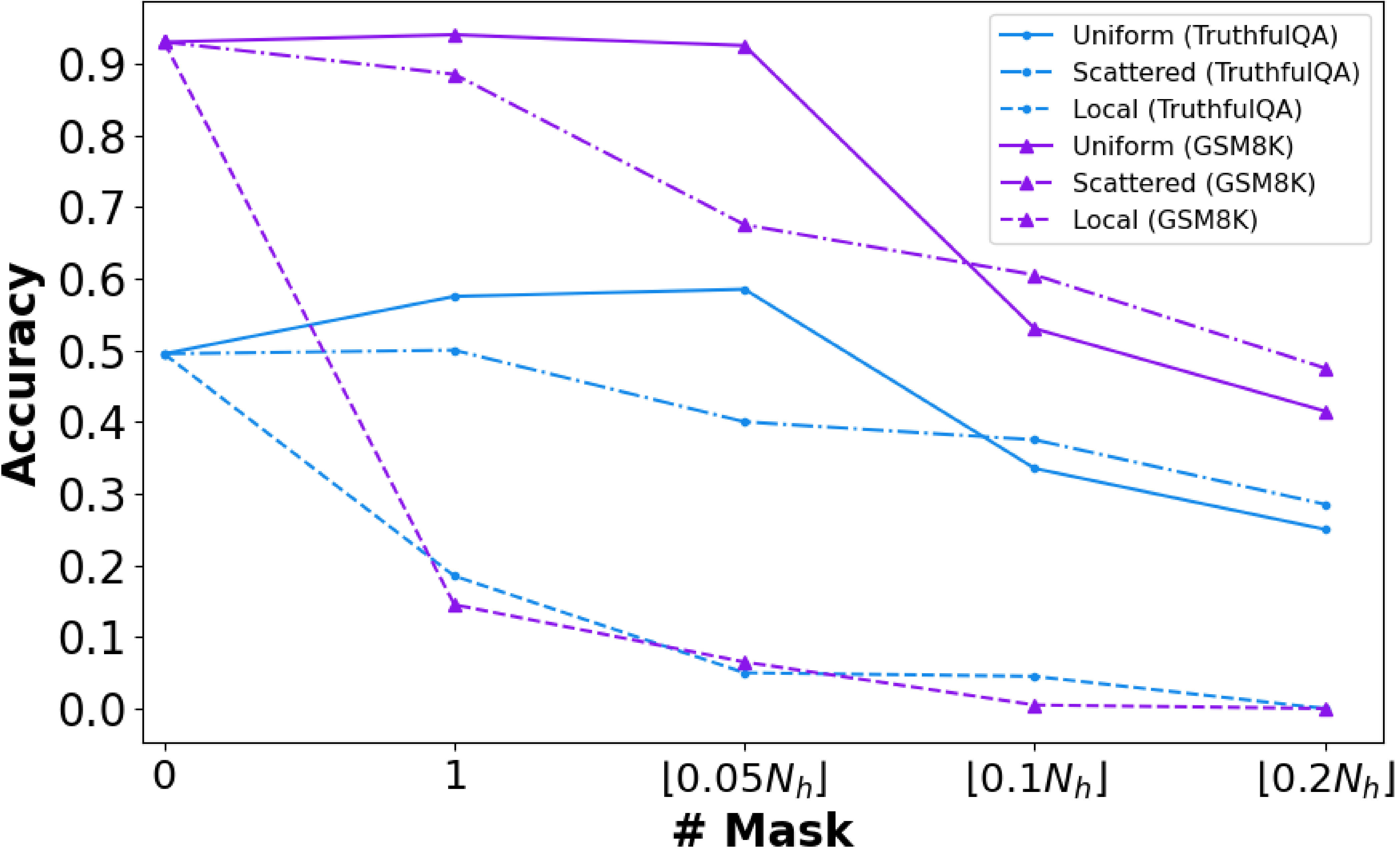}
% \vspace{0.01cm}
\caption{
Accuracy curves of \textit{Qwen2.5-7B-Instruct} evaluated on $\mathcal{D}_\text{ablation}$ of TruthfulQA and GSM8K subsets after masking different amounts of \textit{uniform}, \textit{scattered}, and \textit{local} attention.
}
\vspace{-0.4cm}
\label{fig:attention-mask}
\end{figure}

\subsection{Influence on Model Generation}
\label{sec:attention-influence}
% ！！ 用单图展示展示基于qwen2.5和llama3.1的实验结果：mask掉 local attn, uniform attn 和 scattered attn 对于模型结果的影响 ！！
% ！！ 单表展示不出来变化趋势，这里用plot来展示结果 ！！
% \lzq{add figures to illustrate the influences on the model generation in terms of masking different patterns.}
% Xiao等人和Yu等人讨论过稀疏注意力下注意力池的作用，但密集注意力下不同注意力模式对模型生成的影响目前还尚不清楚。为此，我们研究这三种注意力模式对大语言模型下游任务准确性的影响。我们在5个涵盖了多种推理场景的数据集上进行预实验。
Xiao et al.~\cite{24-streaming-llm} and Yu et al.~\cite{24-act} studied the role of attention sink in sparse attention.
However, the influence of different attention patterns in dense attention is not yet clear.
Therefore, we investigate the effect of the three attention patterns on LLMs' accuracy in downstream tasks.
% （借鉴ACT）讲一下设置
% 具体而言，我们分别从 truthful_qa, gsm8k, logi_qa, commonsense_qa, 和 openbook_qa 数据集中各采样200条数据，并使用Qwen2.5-7B-Instruct在抽样数据集上进行测试。
We conduct preliminary experiments on four datasets that cover diverse scenarios.
Specifically, we randomly sample 200 data points each from the TruthfulQA~\cite{22-truthful-qa}, GSM8K~\cite{21-gsm8k}, LogiQA~\cite{20-logi-qa}, and CommonsenseQA~\cite{19-commonsense-qa} datasets (denoted as $\mathcal{D}_\text{ablation}$ in Section~\ref{sec:ablation-studies}) and test them using \textit{Qwen2.5-7B-Instruct}. 
% 我们比较在普通解码以及分别掩码掉不同比例的均匀注意力、局部注意力和散点注意力的情况下模型的准确性。
% We record the model's accuracy when $\left\lfloor10\%*N_h\right\rfloor$ uniform, local, and dispersed attention are masked.
We record the changes in the model's accuracy when different proportions of uniform, local, and dispersed attention are masked.
% （借鉴act）我们可以从表xxx中得到三个观察：(1) 局部注意力对于模型生成至关重要。即使只掩码掉一个局部注意力也会极大损害模型的生成；(2) 均匀注意力和散点注意力具有一定的“冗余性”，少量掩蔽掉这两种模式对模型生成的影响较小，某些情况下有益于模型生成；(3) 当所掩蔽的注意力较少时，均匀注意力的冗余度比散点注意力更高，掩蔽后对模型生成的影响更小。
% From Table~\ref{tab:attention-mask}, we derive three observations: 
From Figure~\ref{fig:attention-mask}, we derive three observations: 
i) The local attention is crucial for model generation. 
Masking even a single local attention significantly impairs the model generation; 
ii) Uniform and scattered attention exhibit a degree of redundancy.
Masking small amounts of these patterns has few influence on model generation, and in some cases, can enhance it. 
iii) When a small enough amount of attention is masked, uniform attention shows a higher redundancy than scattered attention, thus influencing the model generation less when masked.
% 基于上述观察，我们可以考虑利用注意力中存在的冗余性，利用局部注意力替换最冗余的均匀注意力，来提升模型推理时的注意力利用率以增强大语言模型。
Based on these observations, we consider leveraging the redundancy present in LLMs' dense attention by replacing the most redundant uniform attentions with local attentions to improve attention utilization during generation, thereby enhancing LLMs' performance.

\subsection{Attention Replacement Technique}
\label{sec:attention-replacement-techniques}
% 具体介绍我们的方法
% ！！ 给出一段伪代码以说明我们做注意力替换的时候做了哪些方法 ！！
In light of our findings, we propose a lightweight and effective Attention Replacement Technique (ART) to enhance LLMs by intervening the multi-head attention module during the model generation.
% 回忆公式(3)和(4)，假设一共有$N_h$个注意力头，对于第l层的MHA模块$\text{MHA}^{l}$我们可以把它改写成
Assuming there are a total of $N_h$ attention heads, for the multi-head attention module $\text{MHA}^{l}$ in the $l$-th layer, we can rewrite it as
\begin{align}
    \label{eq:additive-mha}
    % \footnotesize
    % \text{MHA}^{l}(\boldsymbol{X}) &= \sum_{h=1}^{N_h}\boldsymbol{A}_{h}^{l}\boldsymbol{X}{W}_{o,v}^{l} \boldsymbol{W}_{o,h}^{l} = \sum_{h=1}^{N_h}\boldsymbol{A}_{h}^{l}f_h^{l}(\boldsymbol{X}).\\
    % \text{MHA}^{l}(\boldsymbol{X}) &= \sum_{h = 1}^{N_h} \boldsymbol{A}_h^{l} \boldsymbol{X}\boldsymbol{W}_{v,h}^{l}\boldsymbol{W}_{o, h}^{l}
    % \notag \\
    % &= \sum_{h = 1}^{N_h} \boldsymbol{A}_h^{l} f_h^{l}(\boldsymbol{X}).
    \text{MHA}^{l}(\boldsymbol{X}) = \sum_{h = 1}^{N_h} \boldsymbol{A}_h^{l} f_h^{l}(\boldsymbol{X}).
\end{align}
% 设 $\mathcal{A}^l=\{ \boldsymbol{A}_{1}^{l}, \boldsymbol{A}_{2}^{l}, \dots, \boldsymbol{A}_{N_h}^{l}\}$ 表示所有注意力，那么按照公式(7)可以计算出对应的m指标 $\boldsymbol{m}^l=\{m_{1}^{l}, m_{2}^{l}, \dots, m_{N_h}^{l}\}$。将 $\mathcal{A}^l$ 按照 $\boldsymbol{m}^l$ 进行升序排序，得到 $\mathcal{A}^l_\text{sorted}$。
Let $\mathcal{A}^l = \{ \boldsymbol{A}_{1}^{l}, \boldsymbol{A}_{2}^{l}, \dots, \boldsymbol{A}_{N_h}^{l} \}$ represent all the attentions. 
According to equation~(\ref{eq:m-index}), we can compute the corresponding metrics $\boldsymbol{m}^l = \{ m_{1}^{l}, m_{2}^{l}, \dots, m_{N_h}^{l} \}$.
By sorting $\mathcal{A}^l$ in ascending order based on $\boldsymbol{m}^l$, we obtain $\mathcal{A}^l_\text{sorted}$.
% 我们定义k以表示我们将m指标最小的前k个注意力$\mathcal{A}^l_\text{sorted}[:k]$视为均匀注意力并将m指标最大的后k个注意力$\mathcal{A}^l_\text{sorted}[-k:]$视为局部注意力。
We define $k$ to denote that we consider the first $k$ attention heads with the smallest $m$-indices in $\mathcal{A}^l_\text{sorted}[:k]$ as uniform attentions and the last $k$ attention heads with the largest $m$-indices in $\mathcal{A}^l_\text{sorted}[-k:]$ as local attentions.
% 我们用下面两种不同的计算方法来定义$\boldsymbol{A}^{l}_\text{local}$：A^l_h=AAA 
We then define $\boldsymbol{A}^{l}_\text{local}$ as
\begin{align}
    \label{eq:max}
    \textit{Max: } \boldsymbol{A}^{l}_\text{local} &= \mathcal{A}^l_\text{sorted}[-1]\quad \text{or} \\
    \label{eq:mean}
    \textit{Mean: } \boldsymbol{A}^{l}_\text{local} &= \frac{1}{k}\sum_{j=1}^{k}\mathcal{A}^l_\text{sorted}[-j].
\end{align}
% Max表示取m指数最大的局部注意力作为$\boldsymbol{A}^{l}_\text{local}$，而Mean表示取所有局部注意力的平均值来计算$\boldsymbol{A}^{l}_\text{local}$。
\textit{Max} indicates the local attention is selected with the maximum $m$-index as $\boldsymbol{A}^{l}_\text{local}$, whereas \textit{Mean} calculates $\boldsymbol{A}^{l}_\text{local}$ by averaging all local attentions.
% 如果选择(9)计算Alh，那么记作 ART-max, 否则若选择(10)实现Alh，则记作ART-mean.
If $\boldsymbol{A}^{l}_\text{local}$ is calculated through equation~(\ref{eq:max}), ART is specified as ART-max. Otherwise, ART is denoted as ART-mean.
% 我们将$\mathcal{A}^l_\text{sorted}[:k]$替换为$\boldsymbol{A}^{l}_\text{local}$以增强模型的注意力利用率。所以，式(8)可以改写成
We replace $\mathcal{A}^l_\text{sorted}[:k]$ with $\boldsymbol{A}^{l}_\text{local}$ to improve LLMs' attention utilization. Therefore, equation~(\ref{eq:additive-mha}) can be rewritten as 
\begin{align}
    \label{eq:art}
    % \footnotesize
    \text{ART-MHA}^{l}(\boldsymbol{X})  = &\sum_{i\in\mathcal{I}}\boldsymbol{A}^{l}_\text{local}f_{i}^{l}(\boldsymbol{X}) \notag \\
    &+ \sum_{i\notin\mathcal{I}}\boldsymbol{A}_{i}^{l}f_{i}^{l}(\boldsymbol{X}),
\end{align}
% $\mathcal{I}$ 为$\mathcal{A}^l_\text{sorted}[:k]$所对应的编号集合。
where $\mathcal{I}$ is the index set corresponding to $\mathcal{A}^l_\text{sorted}[: k]$.
% 我们在模型生成时将式11应用到模型解码中，(实现效果)
% 我们将公式(11)应用到模型解码中以实现轻量化的注意力干预。
The operation $\text{ART-MHA}^{l}$ is then applied to model inference for efficient intervention.
The full pipeline is demonstrated in Algorithm~\ref{algo:art}.

\section{Experiment}
\label{sec:experiment}

\subsection{Datasets and Models}
\textbf{Datasets}. We evaluate ART on three datasets: 
(i) TruthfulQA~\cite{22-truthful-qa} measures whether LLMs are truthful in responding to questions. 
(ii) LogiQA~\cite{20-logi-qa}, consisting of expert-written questions, aims to test logical reasoning. 
(iii) GSM8K~\cite{21-gsm8k}, containing high-quality elementary school math word problems, is targeted at measuring LLMs' mathematical abilities.
% (iv) CommonsenseQA~\cite{19-commonsense-qa} is a challenging dataset for evaluating LLMs' abilities of commonsense reasoning.
% (v) OpenBookQA~\cite{18-openbook-qa} is another commonsense reasoning dataset in the form of open-book exams.
These datasets cover multiple scenarios for the application of LLMs and are suitable for comprehensive evaluation of their capabilities.

\textbf{Models}. ART is evaluated on prevalent open-source LLMs. They are Llama2-7B-Chat~\cite{23-llama2}, Llama3.1-8B-Instruct~\cite{24-llama3}, Ministral-8B-Instruct-2410\footnote{\href{https://mistral.ai/news/ministraux/}{Ministral blog post}.}, {Qwen2-7B-Instruct}~\cite{24-qwen2.5}, and {Qwen2.5-7B/14B/32B-Instruct}~\cite{24-qwen2.5}.

\subsection{Baselines and Metrics}
% 直接说baseline是什么和metrics是什么。
Following a similar setup~\cite{23-iti,24-act}, we compare ART with the vanilla decoding baseline under the zero-shot Chain-of-Thought~\cite{22-cot} setting. 
We utilize accuracy as the metric for the evaluation of performance.
Notably, questions from GSM8K ask models to answer without answer options available, while questions from other datasets are presented as Multi-Choice Questions (MCQs).

\subsection{Implementation Details}
\label{sec:implementation-details}
We run all our experiments on 1 NVIDIA A800 GPU with 80GB memory.
For TruthfulQA, we directly adopt the whole test dataset for evaluation. For other datasets, we uniformly sample 1,000 from each to build the test datasets $\mathcal{D}_\text{test}$, thus balancing the experimental cost and effectiveness.
All experiments are conducted in a zero-shot CoT setting.
Detailed prompt templates are available in our \href{https://anonymous.4open.science/r/ACL_Attention_Replacement_Techinique}{repository}.
In all our experiments, unless otherwise specified, we use $k=\left \lfloor 0.1*N_h\right \rfloor$, which is derived through the discussion in Section~\ref{sec:ablation-studies}.

\begin{table*}[!htb]
    \renewcommand\arraystretch{1.0}
    \centering
    \scalebox{0.6}{
        \begin{tabular}{l|c|ccccccc}
            \toprule
            \rowcolor{gray!20}
            & & \multicolumn{6}{c}{\textbf{Dataset Performance (\%)}} & \\
            % \cmidrule(lr){3-8}
            \rowcolor{gray!20}
            \multirow{-2}{*}{\textbf{Model}} & \multirow{-2}{*}{\textbf{Method}} & \textbf{TruthfulQA} & \textbf{LogiQA} & \textbf{CommonsenseQA} & \textbf{OpenBookQA} & \textbf{GSM8K} & \textbf{Avg.} \\
            \midrule
            \multirow{3}{*}{Llama2-7B-Chat}
            & Vanilla & 19.0 & 28.4 & 50.8 & 45.9 & 27.2 & 34.3 \\ 
            & ART-max & 21.7 (\textcolor{blue}{+2.7}) & 29.7 (\textcolor{blue}{+1.3}) & 49.5 (\textcolor{blue}{-1.3}) & 44.9 (\textcolor{blue}{-1.0}) & 28.3 (\textcolor{blue}{+1.1}) & 34.8 (\textcolor{blue}{+0.5}) \\ 
            & ART-mean & 21.5 (\textcolor{blue}{+2.5}) & 30.1 (\textcolor{blue}{+1.7}) & 49.6 (\textcolor{blue}{-1.2}) & 45.2 (\textcolor{blue}{-0.7}) & 28.0 (\textcolor{blue}{+0.8}) & 34.9 (\textcolor{blue}{+0.6}) \\ 
            \midrule
            \multirow{3}{*}{Llama3.1-8B-Instruct}
            & Vanilla & 46.5 & 38.8 & 75.5 & 84.4 & 88.1 & 66.7\\ 
            & ART-max & 46.3 (\textcolor{blue}{-0.2}) & 39.6 (\textcolor{blue}{+0.8}) & 76.4 (\textcolor{blue}{+0.9}) & 85.4 (\textcolor{blue}{+1.0}) & 89.2 (\textcolor{blue}{+1.1}) & 67.4 (\textcolor{blue}{+0.7}) \\ 
            & ART-mean & 46.8 (\textcolor{blue}{+0.3}) & 40.2 (\textcolor{blue}{+1.4}) & 76.0 (\textcolor{blue}{+0.5}) & 85.2 (\textcolor{blue}{+0.8}) & 89.0 (\textcolor{blue}{+0.9}) & 67.4 (\textcolor{blue}{+0.7})\\ 
            \midrule
            \multirow{3}{*}{Ministral-8B-Instruct-2410}
            & Vanilla & 46.1 & 41.1 & 71.4 & 84.3 & 89.8 & 66.5\\ 
            & ART-max & 46.4 (\textcolor{blue}{+0.3}) & 44.0 (\textcolor{blue}{+2.9}) & 73.6 (\textcolor{blue}{+2.2}) & 84.1 (\textcolor{blue}{-0.2}) & 90.2 (\textcolor{blue}{+0.4}) & 67.6 (\textcolor{blue}{+1.1}) \\ 
            & ART-mean & 46.2 (\textcolor{blue}{+0.1}) & 43.5 (\textcolor{blue}{+2.4}) & 74.3 (\textcolor{blue}{+2.9}) & 84.3 (\textcolor{blue}{+0.0}) & 90.0 (\textcolor{blue}{+0.2}) & 67.7 (\textcolor{blue}{+1.2})\\ 
          \midrule
            \multirow{3}{*}{Qwen2-7B-Instruct}
            & Vanilla & 41.7 & 42.3 & 73.6 & 85.1 & 88.0 & 66.1 \\ 
            & ART-max & 43.2 (\textcolor{blue}{+1.5}) & 45.0 (\textcolor{blue}{+2.7}) & 73.5 (\textcolor{blue}{-0.1}) & 84.6 (\textcolor{blue}{-0.5}) & 88.9 (\textcolor{blue}{+0.9}) & 67.0 (\textcolor{blue}{+0.9}) \\ 
            & ART-mean & 43.8 (\textcolor{blue}{+2.1}) & 45.5 (\textcolor{blue}{+3.2}) & 73.3 (\textcolor{blue}{-0.3}) & 84.8 (\textcolor{blue}{-0.3}) & 88.8 (\textcolor{blue}{+0.8}) & 67.2 (\textcolor{blue}{+1.1})\\ 
            \midrule
            \multirow{3}{*}{Qwen2.5-7B-Instruct}
            & Vanilla & 51.2 & 50.7 & 78.1 & 87.2 & 93.2 & 72.1\\ 
            & ART-max & 54.6 (\textcolor{blue}{+3.4}) & 52.1 (\textcolor{blue}{+1.4}) & 79.6 (\textcolor{blue}{+1.5}) & 87.6 (\textcolor{blue}{+0.4}) & 93.8 (\textcolor{blue}{+0.6}) & 73.5 (\textcolor{blue}{+1.4}) \\ 
            & ART-mean & 54.0 (\textcolor{blue}{+2.8}) & 52.5 (\textcolor{blue}{+1.8}) & 79.2 (\textcolor{blue}{+1.1}) & 87.2 (\textcolor{blue}{+0.0}) & 93.5 (\textcolor{blue}{+0.3}) & 73.3 (\textcolor{blue}{+1.2}) \\
            \midrule
            \multirow{3}{*}{Qwen2.5-14B-Instruct}
            & Vanilla & 62.3 & 56.9 & 81.5 & 93.0 & 94.3 & 77.6\\ 
            & ART-max & 63.1 (\textcolor{blue}{+0.8}) & 57.8 (\textcolor{blue}{+0.9}) & 82.4 (\textcolor{blue}{+0.9}) & 92.8 (\textcolor{blue}{-0.2}) & 94.8 (\textcolor{blue}{+0.5}) & 78.1 (\textcolor{blue}{+0.5})\\ 
            & ART-mean & 63.3 (\textcolor{blue}{+1.0}) & 57.6 (\textcolor{blue}{+0.7}) & 82.0 (\textcolor{blue}{+0.5}) & 93.0 (\textcolor{blue}{+0.0}) & 94.8 (\textcolor{blue}{+0.5}) & 78.1 (\textcolor{blue}{+0.5}) \\
            \midrule
            \multirow{3}{*}{Qwen2.5-32B-Instruct}
            & Vanilla & 71.6 & 63.0 & 84.6 & 95.0 & 94.7 & 81.8\\ 
            & ART-max & 72.0 (\textcolor{blue}{+0.4}) & 63.2 (\textcolor{blue}{+0.2}) & 84.9 (\textcolor{blue}{+0.3}) & 95.2 (\textcolor{blue}{+0.2}) & 95.4 (\textcolor{blue}{+0.7}) & 82.1 (\textcolor{blue}{+0.3}) \\ 
            & ART-mean & 71.8 (\textcolor{blue}{+0.2}) & 63.5 (\textcolor{blue}{+0.5}) & 84.8 (\textcolor{blue}{+0.2}) & 95.1 (\textcolor{blue}{+0.1}) & 95.5 (\textcolor{blue}{+0.8}) & 82.1 (\textcolor{blue}{+0.3}) \\
            \bottomrule
        \end{tabular}
    }
    \small\caption{Extended performance comparison between ART and vanilla decoding across six reasoning benchmarks. Experiments are conducted over $\mathcal{D}_\text{test}$.}
    \label{tab:main-result}
\end{table*}

\subsection{Main Results}
\label{sec:main-results}
% 表1展示了在所有数据集上
Table~\ref{tab:main-result} presents the experimental results that we validate ART on previously mentioned datasets that cover multiple LLM application scenarios.
% 观察：
% ART对不同模型在不同任务类型的数据集上带来的增益效果不同。
The effects of ART on different models across various task types differ. 
% 总体上看，大部分情况下ART能为大语言模型提升0.6%~1.3%的精确度，某些情况下能实现超过3%的性能提升。例如，当对Qwen2-7B应用ART-mean时，在 LogiQA 上能取得 3.2% 的提升；而当对Qwen2.5-7B应用ART-max时，在 TruthfulQA 上能也取得 3.4% 的提升。
In general, ART can improve the accuracy of LLMs by 0.6\% to 1.7\% in most cases, and in some cases it can achieve performance improvements exceeding 3\%. 
For example, applying ART-mean to \textit{Qwen2-7B-Instruct} results in a 3.2\% improvement in LogiQA.
Similarly, applying ART-max to \textit{Qwen2.5-7B-Instruct} achieves a 3.4\% improvement in TruthfulQA. 
% 从任务类型上来看，相对于常识问答数任务， ART对模型真实性、逻辑推理和数学推理的任务具有更好的增益效果。ART在 CommonsenseQA 和 OpenBookQA 这两类任务上的总体提升幅度平均为0.3%。 对于GSM8K，ART平均能提高0.7%的精准度。进一步，对于 TruthfulQA 和 LogiQA 而言，ART普遍能为LLMs带来1.1%以上的提升。值得注意的是，ART既不需要进行微调，也不需要使用域内数据去确定超参数或构建辅助模型。因此，ART对于LLMs而言带来的性能提升效果是相当可观的。
% In terms of task types, compared to common sense QA, ART exhibits better enhancement in model truthfulness, logical reasoning, and mathematical reasoning. xx
In terms of task types, compared to mathematical reasoning, ART exhibits better enhancement in model truthfulness and logical reasoning.
% The overall improvement for the CommonsenseQA and OpenBookQA tasks is averaged 0.3\%. xx
% For GSM8K, ART can enhance accuracy by an average of 0.7\%.
The overall improvement for the GSM8K task is averaged 0.7\%.
Furthermore, for TruthfulQA and LogiQA, ART generally provides over a 1.1\% improvement for LLMs. 
Notably, ART neither requires fine-tuning nor utilises in-domain data to determine hyperparameters or construct auxiliary models. 
Therefore, the performance gains ART provides to LLMs are quite substantial. 
% 从模型规模上来看，相对于大小为14B/32B模型，ART在7B/8B上带来的性能提升更明显。ART对 qwen2.5-14b/32b 的提升在5个任务上的精确度在0.5%左右，而对 ministral、qwen2 和 qwen2.5 的提升则超过 1.1%。
Regarding model size, ART offers more accuracy enhancement to 7B/8B models than 14B/32B models. 
ART enhances \textit{Qwen2.5-14B/32B-Instruct} models' accuracy in three tasks by approximately 0.7\%, while it improves \textit{Ministral-8B-Instruct-2410}, \textit{Qwen2-7B-Instruct}, and \textit{Qwen2.5-7B-Instruct} models by more than 1.1\%. 
For the ART variants, the ART-max and ART-mean operations exhibit some differences in their calculation of $\boldsymbol{A}^{l}_\text{local}$. 
However, the experimental results indicate that their effect on model generation is almost identical. 
% 对于ART本身而言，ART-max和ART-mean这两种操作方法对于$\boldsymbol{A}^{l}_\text{local}$的计算方式有一定的差异，但实验结果表明，他们对于模型生成的效果影响是几乎一致的。在实际应用中，我们可以酌情选择合适的操作。在后续的实验中，除非特别声明，我们默认以 ART-mean 的方式实现ART。
In practice, we can choose the appropriate operation in a flexible way. 
In subsequent experiments, unless otherwise specified, we default to implementing ART as ART-mean. 
% 表1的结果综合说明，ART对模型带来的增益是稳健的，我们可以将ART应用到多类任务场景上以增强大语言模型。
Table~\ref{tab:main-result} shows that ART provides robust improvement gains in LLMs, suggesting that ART can be applied in multiple tasks to enhance LLMs.

\subsection{Ablation Studies}
% ART的应用涉及到超参数$k$的选择和注意力头的选择与替换等因素。
% 为了探索这些因素对ART的影响，我们从各数据集的训练集中随机采样200条样本以构建$\mathcal{D}_\text{ablation}$并在它上面测试不同的设定下ART的应用效果，以便我们更好地理解和使用ART。
Practicing ART involves factors such as determining hyperparameter $k$ and choice of attention heads, etc.
To explore the effect of these factors on ART, we randomly sample 200 examples from the training set of each dataset to construct $\mathcal{D}_\text{ablation}$. 
We test ART in different settings on $\mathcal{D}_\text{ablation}$ to better understand and utilize it.

\begin{figure}[h]
\centering % 图片居中
\captionsetup{aboveskip=0pt, belowskip=0pt} % 仅对该figure的caption设置前后间距
\includegraphics[width=1.0\linewidth,page=1]{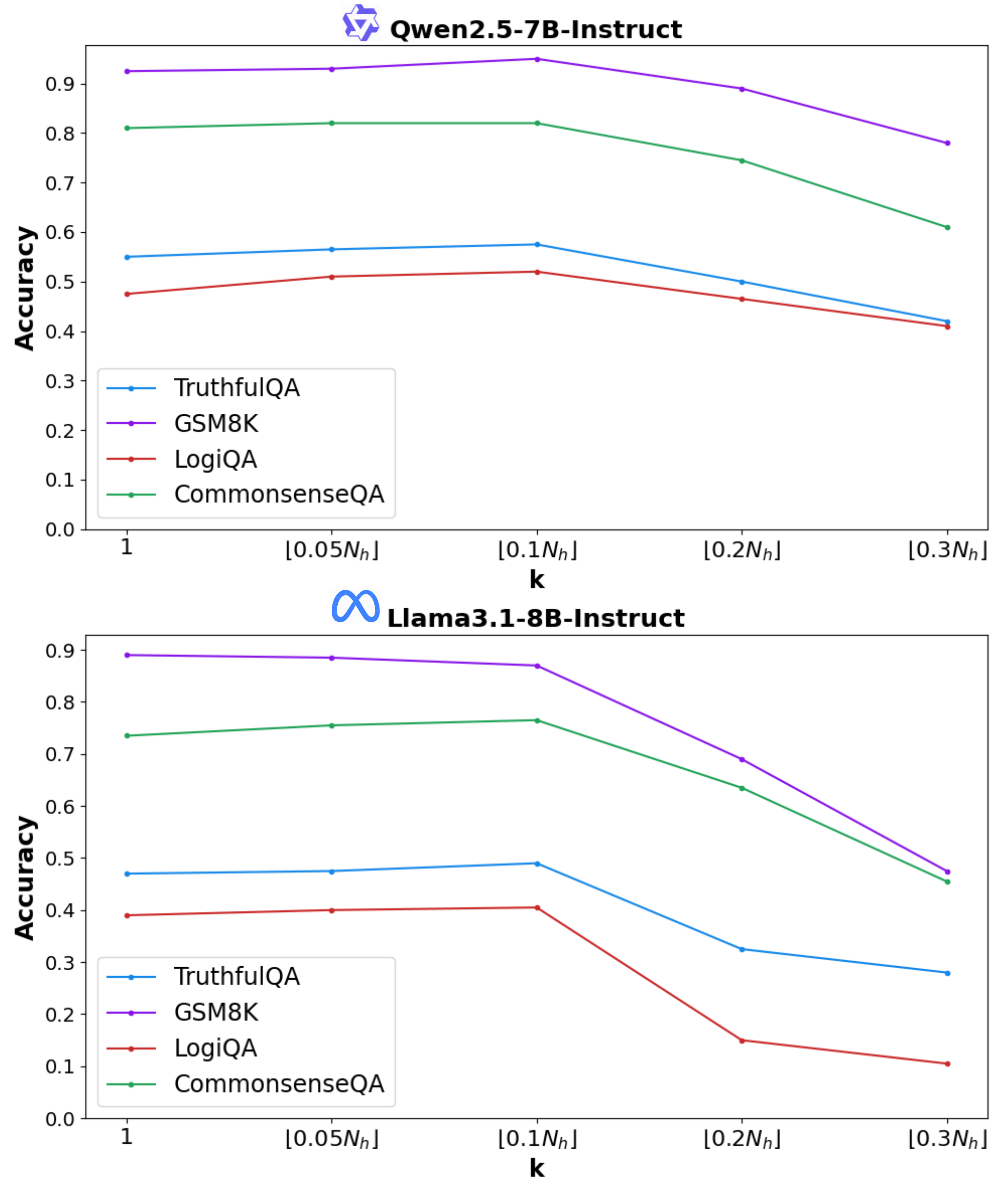}
% \vspace{0.01cm}
\caption{
Accuracy curves of \textit{Qwen2.5-7B-Instruct} and \textit{Llama3.1-8B-Instruct} evaluated on $\mathcal{D}_\text{ablation}$ over different $k$ values.
}
\label{fig:k-selection}
\vspace{-0.4cm}
\end{figure}

\label{sec:ablation-studies}
\textbf{$\boldsymbol{k}$ selection}.
% 我们之前在第三节提到，我们设置$k=\left \lfloor 0.1*N_h\right \rfloor$以进行实验。
% 事实上，k的选择影响着ART的效果。为此，我们测试当$k$从集合$\{0, 1, \left \lfloor 5\%*N_h\right \rfloor, \left \lfloor 10\%*N_h\right \rfloor, \left \lfloor 20\%*N_h\right \rfloor, \left \lfloor 30\%*N_h\right \rfloor\}$里取不同值时，LLMs在$\mathcal{D}_\text{ablation}$上的准确度变化情况。
As mentioned in Section~\ref{sec:implementation-details}, we set $k = \left \lfloor 0.1 * N_h \right \rfloor$ for our experiments. 
In fact, the selection of $k$ significantly impacts the effectiveness of ART. 
Therefore, we investigate the variation in the accuracy of LLMs in $\mathcal{D}_\text{ablation}$ when $k$ takes different values from the set $\{0, 1, \left \lfloor 5\% * N_h \right \rfloor, \left \lfloor 10\% * N_h \right \rfloor, \left \lfloor 20\% * N_h \right \rfloor$, $\left \lfloor 30\% * N_h \right \rfloor\}$.
% 结果如图xxx所示，在多数情况下，随着$k$不断增大，模型准确度先上升后下降。当 $k=\left \lfloor 0.1*N_h\right \rfloor$ 时，准确度达到最最大值。此后，随着$k$的增大，更多的注意力被识别为均匀注意力并被替换掉，这会导致对模型的上下文理解能力的过分损害进而导致模型的性能出现明显下滑。
% 因此，我们经验上将$k$设置为$\left \lfloor 0.1*N_h\right \rfloor$ 以实现ART的最佳效果
The results, as shown in Figure~\ref{fig:k-selection}, indicate that in most cases, as $k$ increases, the accuracy of the model increases first and then decreases.
When $k = \left \lfloor 0.1 * N_h \right \rfloor$, the accuracy reaches the maximum. 
Beyond this point, further increasing $k$ leads to more uniform attentions being identified and replaced, which can overly impair the model's context understanding ability, thereby resulting in a significant accuracy decrease. 
Consequently, we empirically set $k = \left \lfloor 0.1 * N_h \right \rfloor$ to achieve the optimal effectiveness of ART.

% \begin{table}[!htb]
%     % \vspace{0.2cm}
%     \renewcommand\arraystretch{1.1}
%     \centering
%     \scalebox{0.6}{
%     % \resizebox{\textwidth}{!}{%
%         \begin{tabular}{c|c|ccc}
%             \Xhline{1pt} % \usepackage{booktabs}
%             \textbf{LLM}& \textbf{Method}& \textbf{TruthfulQA}& \textbf{LogiQA}& \textbf{GSM8K} \\ % 占位行
%             \hline
%             \multirow{2}{*}{\rotatebox{0}{Qwen2.5-7B-Instruct} }
%             & \text{ART} & 52.5 & 51.0 & 94.0  \\ 
%             & \text{ART-inverse} & 0.0 & 2.0 & 7.5 \\ 
%             \hline
%             \multirow{2}{*}{\rotatebox{0}{Llama3.1-8B-Instruct} } & \text{ART} & 46.0 & 40.5 & 87.0 \\ 
%             & \text{ART-inverse} & 38.5 & 31.5 & 64.5  \\ 
%             \Xhline{1pt}
%         \end{tabular}
%     % \end{tabularx}
    
%     }
%     % \captionsetup{aboveskip=8pt, belowskip=0pt} % 仅对该figure的caption设置前后间距
%     \small\caption{
%     Performance comparison between ART and ART-inverse. 
%     Tests are conducted on $\mathcal{D}_\text{ablation}$.
%     }
%     \label{tab:replacement-direction}
% \end{table}

\begin{table}[!htb]
    \renewcommand\arraystretch{1.1}
    \centering
    \scalebox{0.62}{
        \begin{tabular}{lccccc}
        \rowcolor{gray!20}
            \toprule
            \textbf{LLM} & \textbf{Method} & \textbf{TruthfulQA} & \textbf{LogiQA} & \textbf{GSM8K} \\
            \midrule
            \multirow{4}{*}{Qwen2.5-7B-Inst} & ART & 52.5 & 51.0 & 94.0 \\ 
             & \text{Vanilla} & 51.2 & 50.7 & 93.2  \\ 
            & ART-inverse & 0.0 & 2.0 & 7.5 \\ 
             & \text{ART-scattered} & 53.2 & 51.7 & 93.5 \\ 
            
            \midrule
            \multirow{4}{*}{Llama3.1-8B-Inst} & ART & 46.0 & 40.5 & 87.0 \\ 
            &\text{Vanilla} & 46.5 & 38.8 & 88.1 \\ 
            & ART-inverse & 38.5 & 31.5 & 64.5  \\ 
            & \text{ART-scattered} & 46.6 & 39.3 & 89.2  \\ 
            \bottomrule
        \end{tabular}
    }
    \small\caption{
    Performance comparison between ART and ART-inverse. 
    Tests are conducted on $\mathcal{D}_\text{ablation}$.
    }
    \label{tab:replacement-direction}
    \vspace{-0.2cm}
\end{table}

\textbf{Replacement direction}.
Section~\ref{sec:attention-influence}.
% 在章节3的讨论中我们知道，局部注意力对模型生成十分重要，掩蔽局部注意力会使模型无法完成任务。另一个直接的问题是，若局部注意力并非被掩蔽掉，而是被均匀注意力替换掉，模型生成会受到何种影响？为此，我们测试了ART-inverse，它将ART中的替换关系逆转，使用均匀注意力替换局部注意力。测试结果展示在表xxx中。
As discussed in Section~\ref{sec:attention-influence}, local attention is crucial for model generation, and masking local attention could make LLMs fail to complete tasks. 
An additional question arises: what if local attention is not masked but replaced with uniform attention? 
To investigate this, we test \textit{ART-inverse}, which reverses the substitution in ART by replacing local attention with uniform attention. 
% 可以看到，对Qwen2.5-7b-instruct来说，ART使其在下游任务上几乎完全失效。而对Llama3.1-8b-instruct而言，ART使其在下游任务上出现显著的性能下滑。
% 表xx和表xx的结果共同说明了局部注意力对于模型生成至关重要，对局部注意力的调整很可能会严重损害LLMs的性能。
Table~\ref{tab:replacement-direction} conveys that for \textit{Qwen2.5-7B-Instruct}, ART renders it almost entirely ineffective in downstream tasks, while for \textit{Llama3.1-8B-Instruct}, ART leads to a significant performance degradation. 
Results from Table~\ref{tab:main-result} and Table~\ref{tab:replacement-direction} collectively demonstrate that local attention is crucial, and modifications to local attention could severely impair model performances.

% \begin{table}[!htb]
%     % \vspace{0.2cm}
%     \renewcommand\arraystretch{1.1}
%     \centering
%     \scalebox{0.6}{
%     % \resizebox{\textwidth}{!}{%
%         \begin{tabular}{ccccc}
%         \rowcolor{gray!20}
%             \Xhline{1pt} % \usepackage{booktabs}
%             \textbf{LLM}& \textbf{Method}& \textbf{TruthfulQA}& \textbf{LogiQA}& \textbf{GSM8K} \\ % 占位行
%             \hline
%             \multirow{2}{*}{\rotatebox{0}{Qwen2.5-7B-Inst} }
%             & \text{Vanilla} & 51.2 & 50.7 & 93.2  \\ 
%             & \text{ART-scattered} & 53.2 & 51.7 & 93.5 \\ 
%             \hline
%             \multirow{2}{*}{\rotatebox{0}{Llama3.1-8B-Inst} } & \text{Vanilla} & 46.5 & 38.8 & 88.1 \\ 
%             & \text{ART-scattered} & 46.6 & 39.3 & 89.2  \\ 
%             \hline
%             \multirow{2}{*}{\rotatebox{0}{Ministral-8B-Inst} } & \text{Vanilla} & 46.1 & 41.1 & 89.8 \\ 
%             & \text{ART-scattered} & 46.2 & 42.3 & 90.3  \\ 
%             \Xhline{1pt}
%         \end{tabular}
%     % \end{tabularx}
    
%     }
%     % \captionsetup{aboveskip=8pt, belowskip=0pt} % 仅对该figure的caption设置前后间距
%     \small\caption{
%     Performance comparison, conducted on $\mathcal{D}_\text{test}$, between vanilla decoding and ART-scattered. 
%     }
%     \label{tab:replacement-with-scattered-attention}
%     \vspace{-0.2cm}
% \end{table}

% 这个测试将 mask attention 用 scattered attention 来替换会有怎样的效果（相对于local attention）
\textbf{Ways to replace attentions}.
% 在小节V中我们描述了本文所提出的注意力替换方法, ART。其中，$\mathcal{A}^{l}_\text{local}$ 作为替换后目标注意力，由局部注意力计算得来。
In Section~\ref{sec:attention-replacement-techniques}, we describe ART in detail.
Here, $\boldsymbol{A}^{l}_\text{local}$ the target attention, is calculated through local attention.
% 一个直观的想法是是否可以使用散点注意力来计算目标注意力$\mathcal{A}^{l}_\text{local}$以增强模型生成。
A direct question is whether scattered attention can also be utilized to compute the target attention to enhance model generation.
% 因此，我们将该方法记为 ART-scattered 并且在数据集 $\mathcal{D}_\text{test}$ 上测试其效果。
Therefore, we refer to this method as ART-scattered and evaluate its performance on the primal test dataset $\mathcal{D}_\text{test}$.
% 表3的结果表明，ART-scattered 仍可以一定程度上增强LLMs，虽然和ART相比其增益的效果有一定收缩。这一定程度上说明了目标注意力的计算是十分灵活的。我们把如何构建更好的目标注意力留为未来工作。
Results presented in Table~\ref{tab:replacement-direction} indicate that ART-scattered can still enhance LLMs to a certain extent, although its effectiveness has decreased somewhat compared to ART. 
This suggests that the choice of target attention is quite flexible. 
We leave the exploration of constructing better target attention for future work.

\textbf{Comparison to other inference intervention methods}.
% 我们将ART与波束搜索这种在大模型应用中广泛使用的解码策略进行对比，并且我们也测试将ART与波束搜索结合的方法(ART-Beam)以辨别ART是否可以通过与波束搜索结合实现进一步增强大语言模型。
We compare ART with \textit{Beam Search}, a widely used decoding strategy in LLM applications. 
Additionally, we evaluate a hybrid approach that combines ART with beam search, called \textit{ART-Beam}, to investigate whether ART can further enhance LLMs. 
% 同时，我们还将ART与ITI,ACT和DoLa三种同属在模型推理时进行干预方法进行比较。ITI利用域内训练数据为大模型内的所有注意力头训练特定的探针，以便判断大模型内的真实头并在推理时对他们进行编辑。ACT利用训练数据识别大模型内部适合进行注意力矫正的注意力头，并在后续模型推理中对这些注意力头内的注意力模式进行编辑。DoLa则主要对比模型内由不同层得到的逻辑值之间差别来增强模型生成的真实性。
Furthermore, we compare ART with three other intervention-based methods, ITI~\cite{23-iti}, ACT~\cite{24-act}, and DoLa~\cite{24-dola}.
ITI uses in-domain training data to train probes for every attention head in the language model to identify the \textit{Truthfulness Head}, which are then edited during inference.
ACT leverages training data to identify attention heads suitable for attention calibration and subsequently edits these attention heads on the fly during inference. 
DoLa enhances the model truthfulness by contrasting the differences in logits obtained from final layers versus premature layers.
% 结果展示在表4中。可以看出，在多数任务上，ART能够取得和ITI以及ACT相当的、甚至更好的表现。而相对于ITI和ACT，ART不需要使用下游任务相关的数据进行训练或者预实验。仅需通过对模型浅层进行简单的注意力替换即可增强模型性能，这使得ART具有更好的可应用性。
% The results in Table~\ref{tab:baselines} show that ART achieves comparable or even superior performance to ITI, ACT and DoLa on most tasks. 
Table~\ref{tab:baselines} shows that ART can be combined with beam search for further improvement and achieves comparable or even superior performance to ITI, ACT, and DoLa on most tasks. 
However, unlike ITI and ACT, ART does not require task-specific downstream data for training or pilot experiments. 
Instead, it enhances LLMs by simply replacing attention patterns in the shallow layers, making ART more practical.

\begin{table}[t]
    \renewcommand\arraystretch{1.0}
    \centering
    \scalebox{0.70}{
        \begin{tabular}{c|cccc}
        \rowcolor{gray!20}
            \toprule
            \textbf{Method} & \textbf{Method} & \textbf{TruthfulQA} & \textbf{LogiQA} & \textbf{GSM8K} \\
            \midrule
            \multirow{7}{*}{Llama2-7B} & Vanilla & 19.0 & 28.4  & 27.2 \\ 
            & Beam Search &  21.4 & 29.2 & \uline{29.5} \\  
            & ITI & 19.9 & 28.2  & 28.1 \\ 
            & ACT & 18.7 & \uline{30.3} & 26.7 \\
            & DoLa & 21.5 & 27.8 & 28.7  \\
            & ART-greedy & \uline{21.5} & 30.1 & 28.5 \\ 
            \rowcolor{gray!20}
            & ART-beam & \textbf{22.7} & \textbf{30.5} & \textbf{30.1} \\ 
            \hline
            \multirow{7}{*}{Llama3-8B} & Vanilla & 45.1 & 38.5  & 83.1 \\ 
            & Beam Search &  \uline{46.0} & 41.0 & \uline{84.3} \\  
            & ITI & \uline{46.0} & 37.3  & 66.7 \\ 
            & ACT & 45.2 & 38.0 & 72.8 \\
            & DoLa & 45.3 & 38.8 & 83.2  \\
            & ART-greedy & 45.8 & \uline{41.4} & 84.0 \\ 
            \rowcolor{gray!20}
            & ART-beam & \textbf{46.4} & \textbf{42.7} & \textbf{85.8} \\ 
            \bottomrule
        \end{tabular}
    }
    \small\caption{
    Performance comparison between ART and other inference intervention methods, by evaluating \textit{Llama2-7B-Chat} and \textit{Llama3-8B-Instruct} on $\mathcal{D}_\text{test}$.
    \textbf{Bold} and \uline{underline} denote the best and second best. % respectively.
    }
    \label{tab:baselines}
\end{table}

\section{Conclusion}
\label{sec:conclusion}
% 在这篇论文中，我们可视化大语言模型内不同的注意力模式并且讨论了在模型浅层的三类密集注意力的特点。
In this paper, we visualize different attention patterns within LLMs and discuss the characteristics of three types of dense attention (uniform attention, local attention, and scattered attention) in LLMs' shallow layers. 
% 基于我们的可视化和试点实验，我们发现对模型而言局部注意力的重要性以及均匀注意力的冗余性。
Based on our visualizations and pilot experiments, we discover the significance of local attention and the redundancy of uniform attention for LLMs. 
% 基于我们的发现，我们提出了一种注意力替换技术(ART)，它在模型推理时用局部注意力替换均匀注意力以提高大语言模型在模型推理时的内部注意力利用率以此来增强大语言模型在下游任务上的表现。
In light of our findings, we propose the Attention Replacement Technique, named \textbf{ART}, which replaces uniform attention with local attention during model inference to enhance the internal attention efficiency of LLMs and thus improve their performance on downstream tasks. 
% 多种开源大语言模型在多类数据集上的实验结果验证了ART的有效性。
Experimental results across multiple open-source LLMs on various datasets demonstrate the effectiveness of ART.

% \section*{Limitation}
% It performs an empirical analysis of hallucinations and attention head patterns. However, there may be more interpretable methods that could offer deeper understanding or targeted interventions.
% ART relies on modifying attention patterns in shallow layers, assuming that uniform attention is a primary cause of hallucinations. However, hallucinations may also arise from other factors, such as model architecture, training data biases, or decoding strategies. ART does not address these potential sources of error. While ART's training-free nature is a strength, it may also limit its ability to adapt to specific tasks or domains. 

% \section*{Acknowledgments}
% not for now

\bibliography{main}

\appendix

\section{Algorihtm Illustration}
\label{sec:appendix_algo}

Algorithm~\ref{algo:art} fully depicts the whole pipeline of integrating ART into the model inference.
$k$ refers to the size of the number of identified uniform attention and local attention, which is set to $k = \left \lfloor 0.1 * N_h \right \rfloor$ by default discussed in Section~\ref{sec:ablation-studies}.
$L_s$ indicates the number of layers identified as shallow layers, which is set to $L_s=2$ in this paper.
$L$ represents the total number of layers the language model $\mathcal{M}$ have.
% 参考：https://blog.csdn.net/qq_43760191/article/details/121519247
% algorithm
% algorithmic
% \begin{breakablealgorithm} % 可分页伪代码
\begin{figure}[h]
    \centering
    \begin{minipage}{0.48\textwidth}
    
% \begin{algorithm}[!h] % 不分页伪代码
\begin{algorithm}[H] % 不分页伪代码
    \caption{\textbf{A}ttention \textbf{R}eplacement \textbf{T}echnique}
    \label{algo:art}
    \renewcommand{\algorithmicrequire}{\textbf{Input:}}
    \renewcommand{\algorithmicensure}{\textbf{Output:}}
    \begin{algorithmic}[1]
        \REQUIRE question $\mathcal{Q}$, 
        model $\mathcal{M}$,  
        tokenizer $\mathcal{T}$,
        max tokens $m$, 
        \# uniform/local attention $k$,
        shallow layer $L_s$,
        \# layer $L$,
        %%input
        \ENSURE response $\mathcal{R}$ to question   %%output
        
        \STATE  \texttt{gen\_tokens} $\leftarrow$ [] 
        \FOR{each $i \in [1,m]$}
            \STATE $\mathcal{X}\leftarrow$ \textsc{Encode}($\mathcal{T},\mathcal{M}, \mathcal{Q},$ \texttt{gen\_tokens})
            \STATE \textcolor{gray}{
                \small\textit{\# Forward pass: go through all $L$ layers }
                }
            \FOR{each $l \in [1,L]$}
                \STATE \textcolor{gray}{
                    \small\textit{\# Multi-Head Attention }
                    }
                \IF{$l\le L_s$ } 
                    \STATE \textcolor{gray}{
                    \small\textit{ART}
                    }
                    \STATE $\mathcal{X}\leftarrow \mathcal{X} + \text{ART-MHA}^{l}_{\mathcal{M}, k}(\text{LN}(\mathcal{X}))$
                \ELSE
                    \STATE \textcolor{gray}{
                    \small\textit{ Vanilla MHA}
                    }
                    \STATE $\mathcal{X}\leftarrow \mathcal{X} + \text{MHA}^{l}_{\mathcal{M}}(\text{LN}(\mathcal{X}))$
                \ENDIF
                \STATE \textcolor{gray}{
                    \small\textit{\# Feed-Forward Network }
                    }
                \STATE $\mathcal{X}\leftarrow \mathcal{X} + \text{FFN}^{l}_{\mathcal{M}}(\text{LN}(\mathcal{X}))$
            \ENDFOR
            \STATE $\mathcal{P}\leftarrow$ \textsc{lm\_head}($\mathcal{M}$, $\mathcal{X}$)
            \STATE $t_i\leftarrow$ \textsc{Decode}($\mathcal{T}$, $\mathcal{P}$)
            \STATE \texttt{gen\_tokens.append}$(t_i)$
            \IF{stopping criteria satisfied}
                \STATE \texttt{break}
            \ENDIF
        \ENDFOR
        
        \RETURN \textsc{tokens\_to\_text}($\mathcal{T}$, \texttt{gen\_tokens}) 
    \end{algorithmic}
\end{algorithm} 
% \end{breakablealgorithm}

\end{minipage}
\end{figure}

As descried in Algorithm~\ref{algo:art}, ART is only applied to the multi-head attention module of $\mathcal{M}$ and activated when the current layer is shallow.
When ART is not activated, $\mathcal{M}$ performs vanilla autoregressive decoding.

\end{document}